\documentclass[journal]{IEEEtran}
\ifCLASSINFOpdf
\else
\fi
\usepackage{enumitem}
\usepackage{mathtools} 
\usepackage{amssymb} 
\usepackage{amsmath}
\usepackage{tabularx}
\usepackage[font=footnotesize,labelfont=bf]{caption}
\usepackage{url}
\usepackage{cite}
\usepackage[pdftex]{graphicx}

\begin{document}
%
\title{Continuous and Simultaneous Gesture and Posture Recognition for Commanding a Robotic Wheelchair; Towards Spotting the Signal Patterns}
%
%
%

\author{Ali~Boyali, 
        Naohisa~Hashimoto,  Manolya~Kavakli 
\thanks{The first two authors are with the National Institute of Advanced Industrial Science and Technology (AIST), Tsukuba, Ibaraki, 305-8561, Japan. The third author is the head of Virtual Reality research group at the Department of Computing, Macquarie University. 
E-mails: \{ali-boyari\}  \{naohisa-hashimoto\}@aist.go.jp and manolya.kavakli@mq.edu.au}    

\thanks{Manuscript received April 19, 2005; revised December 27, 2012.}}

%
%

\markboth{Journal of \LaTeX\ Class Files,~Vol.~11, No.~4, December~2012}%
{Shell \MakeLowercase{\textit{et al.}}: Bare Demo of IEEEtran.cls for Journals}
%



\maketitle

\begin{abstract}
Spotting signal patterns with varying lengths has been still an open problem in the literature. In this study, we describe a signal pattern recognition approach for continuous and simultaneous classification of a tracked hand's posture and gestures and map them to steering commands for control of a robotic wheelchair. The developed methodology not only affords  100\% recognition accuracy on a streaming signal for continuous recognition, but also brings about a new perspective for building a training dictionary which eliminates human intervention to spot the gesture or postures on a training signal. In the training phase we employ a state of art subspace clustering method to find the most representative state samples. The recognition and training framework reveal boundaries of the patterns on the streaming signal with a successive decision tree structure intrinsically. We make use of the Collaborative ans Block Sparse Representation based classification methods for continuous gesture and posture recognition.  
\end{abstract}


%
\IEEEpeerreviewmaketitle

\section{Introduction}
%
%
%
%
\IEEEPARstart{T}{he} advancement in medicine and technology has led a drastic change in the age composition of the world's population. The decline in the fertility and decreasing mortality rates create a lopsided age distribution  with and increase in the elderly proportion. The demographic figures, dependency; the ratio of the dependent ages to the number of working-ages and old-age support ratio; the ratio of the dependent elderly to the working age which invoke fiscal burden to the societies, are the two immediate impacts to the well-being of the public and social services. 

As stated in the United Nations (UN) Report \cite{unhr}, the elderly population over 60 was approximately 841 million, corresponding to 11.1\% of the overall world's population in 2013, which is four-times of the population in 1950. The report also projects that, by 2050, the elderly population will triple and exceed two billion level nearly doubling the proportion of the elderly 21.1\%. Japan is the highest rate of the elderly in the world. The ratio of the elderly over 60 years age is reported as 33\% in August 2014 \cite{jpstat}.

The health conditions deteriorates and the prevalence rate of the chronic diseases increases by aging processes. These conditions as well as the decrease in the psycho-motor skills make the elderly dependent with some forms of disabilities which lead to slow movement even immobility in some period of older ages. Because of the aging population and the nature of aging, the portion of the elderly to the people with moderate and severe disability is the highest among the age bandwidths in the world which was estimated as 46.1\% in 2004 \cite{gbid}. 

The homebound and socially isolated individuals are more likely to get trapped into the depression disorders which cause poor life quality and satisfaction. Besides, the contribution of the older ages to the economy is adversely affected. Likewise, the mobility impairment and being psychically limited may cause frustration in the childhood age ranges, and may create co-morbid cognitive and emotional disabilities during psychological development. Children learn by interaction with their peers and exploring the environment. Enabling the children and elderly with ambulatory impairments, or severe disabilities to have independent functional mobility is on the top list of the United Nations (UN) Convention on the Rights of Persons with Disabilities (CRPD) \cite{notes2007convention}.

Although, no normative definition has yet existed, the UN CRPD brings different definition of the disability cogently in terms of interaction of the people with disabilities with their environment. According to the convention, disability is not incapability per se, but it is created by the non-accessible environments and lack of the proper  assistive devices. Hence, the definition posits the disability as a human right issue. Insufficient number of the care-giving and nursing personnels due to the increasing dependency and old-age support ratios, exorbitant public expenditures on the social and health services call upon the states to take urgent measures to create more-accessible, disabled friendly city and building structures and the public and private institutions to create more smart assistive devices. 

As the rapid developments in the computer and sensor technologies which are embedded in virtually every mainstream products, mobile phones, tablets even electronic children toys, the power wheelchairs have become  more affordable devices, rather obsolete and off-the-shelf \cite{bostelman2006survey} due to emergence of more complicated robotic wheelchairs. 
 
In the UN report, the portion of the wheelchair users in the world population with moderate and severe disability is given as 10\% and the report suggest as a rule of thumb that, the population of the wheelchair users corresponds to 1 \% of a country's population over 65 million \cite{wheelchairratio}. In Japan, the number of people with moderate to severe disabilities is around 3.5 million. Half of this population have ambulatory impairment  \cite{cabinet}. The number of mobility devices sold in Japan is 640.000 including 120.000 electric wheelchairs in 2013.  

The manufacturers and the application developers have been putting a great deal effort to add more modalities to the Human Machine and Computer Interfaces (HMI and HCI), to the mainstream products such as the mobile phones and tablets to reach a vast market engulfing the technology challenged population. However, when the people with disability or elderly are concerned, the design of HCIs through which these cohort groups can interact with the mobility or assistive robots in an intuitive manner requires even more effort then that of the mainstream products, especially in the health related fields. 

In this study, we explain the development stages of HCIs for commanding a robotic wheelchair in a broad sense while giving comparisons of the concepts, methods and algorithms for gesture and posture classification. Yet, in particular, we introduce a combined gesture and posture recognition method from high dimensional streaming signal and the use of these methods for controlling a robotic wheelchair by hand postures and gestures. We used a Leap Motion sensor for the proof-of-concept system and demonstrations to capture the postures and gestures of a detected hand, however we also tested the implemented algorithms for the signals captured in  various sensor domain such as Wiimote Infra-Red (IR),  and Android motion sensors to classify the signals depending on the aim of application. 

We employ the state of art Sparse and Collaborative Representation based Classification methods (SRC and CRC) \cite{wright2009robust, zhang2011sparse} to develop simultaneous gesture and posture recognition -in general signal classification- algorithms for various sensor domain. This study is a part of an ongoing project by which the authors aim to build an entire context aware system capable of spot gesture and posture signal of an any observed body part of the occupier, measure the physical and physiological factors to detect whether the occupier in a dangerous situation from the seating postures or doing an intentional movement to command the wheelchair.

The contributions of this paper to the signal pattern classification, gesture and posture recognition are four fold;  

\begin{description}
	
	\item [First]  Proposes an approach to spot signal patterns on the fly, and elucidate the application of the proposed method to control a robotic wheelchair by continuous hand postures and gestures.   
	
	\item[Second] Presents a framework for building an effective training dictionary from a streaming signal automatically, eliminating the need of human intervention to put the signal patterns into the classes to be recognized in the dictionary matrices.    
	
	\item [Third] Compares the Block-SRC and CRC methods for classification and recognition of different signal domains and demonstrates the use of these methods on mobile devices.    
	
	\item [Forth] Sheds light on how the proposed methods can be applied to detect residual voluntary muscle activities and body part movements to enable the individuals with severe mobility impairments and disability to command a robotic wheel-chair in an intuitive manner. 
	
\end{description}

The rest of the paper is organized as follows. In Section II, we give a brief review of the interfaces used for power wheelchairs, gesture and posture recognition studies as well as the open issues in intuitive interfaces research fields. In Section III, we detail technical background of the signal pattern classification methods. Section IV is dedicated to implementation, in which system and software architecture as well as the training procedure are presented. We give, compare and discuss simulation results in Section V. Finally, the paper ends with the Conclusion and Future works in Section VI.
  
\section{Related Studies In Literature}

\subsection{Power Wheelchair Control Interfaces}

The goal of the research is to develop methods and algorithms to capture the voluntary body part movements adequately and map them to the sound control commands for a robotic wheelchair navigation. Insomuch as, the target population is the individuals with severe disability or mobility impairments, the proposed algorithm must rely on intuitive movements that reduces the cognitive load of the performer and discriminates the voluntary movements robustly with a high recognition accuracy. Since, many sensors and sensor fusion can be used to capture any voluntary body or muscle activity, the signal domain independence is important as the previously listed user interface requirements.   

The joystick control has been the prevailing steering interface for the power wheelchair. As such, it is the mostly prescribed rehabilitation device by the clinicians. According to the study \cite{fehr2000adequacy}, 9-10\% of the patients with severe disabilities who were subject to power wheelchair training find it quite arduous, and 40\% of the same group state the steering and maneuvering as an extremely challenging or impossible task. In the same study, 32\% of the clinician surveyed from the different states and rehabilitation centers in the USA report that, the number of patients who are fit or not for power wheelchair rehabilitation are equally distributed.

The wheelchair interfaces other than the conventional joystick control can be divided into two categories with respect to what kind of signals are used to command; these are the biosignal and non-biosignal based control interfaces \cite{jose2013development}. The studies in the literature make use of biosignal or bioelectrical activities from different regions and parts of the body to detect and map the subject's intention for control commands. 

Among these studies, electrical activities of the brain, active muscles and eye movements are measured and monitored by Electroencephalogram (EEG) \cite{craig2007adaptive, iturrate2009synchronous, carlson2013brain}, Electromyogram (EMG) \cite{felzer2002hawcos, wei2009use, han2001development} and Electrooculography (EOG) \cite{barea1999guidance, al2011guiding} respectively or combined modalities \cite{hashimoto2009wheelchair} to infer the intention of the wheelchair occupant from the signal patterns. 

The biosignal based interfaces are most suitable for the people with severe and multible disabilities as the only resource can be acquired from them is biosignals  \cite{jose2013development}. Nonetheless, in addition to hindering nature due to the high number of wired electrodes and conductivity gel, requirements of high mental attention that increase cognitive load, the classification of the EEG signals suffer from low level of accuracy and yet, it is still not a suitable interface for the real-world control applications. The EMG and EOG based studies utilize the similar type of obstructive electrodes which brings further restriction to the occupants. The Finite State Machines by limited number of switching commands, statistical features extraction, neural network and support vector machine based classifiers are used to classify and detect the intention of the occupant. The misclassification levels vary between 8-20\% in the biosignal based studies available in the literature. Along with the listed drawbacks, ad-hoc feature extraction and filtering methods  used for fine tuning and customization make the proposed schemes only application specific. We refer the reader to the studies \cite{bashashati2007survey, oskoei2007myoelectric} for comprehensive surveys on the EEG and EMG interfaces.   

The non-biosignal based controllers fall into three categories which are the interfaces that rely on the mechanical excitation,  sip-and-puff, chin and head inputs; optical methods, eye gaze, head and hand tracking; and vocal inputs, voice and speech recognition. The first group suits for more flexible patients, therefore is out of context in this paper. The interfaces rely on the vocal inputs suffer from background noise, whereas the recognition accuracy is highly dependent  on speech dexterity \cite{boyali2014hand}.

The optical tracking based user interface studies which are scarce in the literature use template matching methods to detect the current position of the tracked hand or head \cite{gray2007head}. Although a plethora of gesture and gesture recognition studies for different application domains are accessible in the literature, to best of our knowledge and literature review, these modalities are not common for wheelchair control interfaces. 

\subsection{Gesture and Posture Recognition}
Many scientists from different disciplines have a consensus on that the human communication evolved from the primitive forms of gestures; pointing and pantomiming which then evoked speech and language in human communication during the course of evolution \cite{tomasello2010origins, corballis2003hand}. Having ancestral roots deeper in the mind, the gesture is a visible bodily action that occurs naturally in human communication. As such, it is the most intuitive modality to convey the intention and thoughts during communication. Since gesture can manifest itself in many dimensions such as trajectory, space, rhythm, time \cite{de2009language}, it has been the capstone for development of various user interfaces in various  research areas such as gaming, robotics and virtual reality applications in the last decade.  

When the elderly and children with severe and multiple disabilities whose only resource are the residual muscle activities, are considered, the features of gestures; intuitiveness, diversity and alleviating cognitive load \cite{goldin2001explaining} make the gestures a good candidate for HCI of power wheelchairs. 
 
Wheelchair navigation, as in with joystick control requires a limited number of steering commands; turn left, right, forward, reverse and stop, if requested, increasing and reducing the speed. Designing gesture recognition algorithms using different kinds of state of the art machine learning and classification methods with a high recognition accuracy is trivial, however additional requirements such as, picking the gesture from a signal of continuously observed body part; namely spotting the gesture, building a robust training dictionary; low level requirements such as switching in and out of the navigation mode, rejecting the unintentional movements being recognized as a gesture command, handling the dangerous situations and risk management are of great importance and must be taken into account in designing gestural interfaces in the health related fields.   

Spotting gestures from a streaming single or multi-dimensional signal has been still an open problem in the related research field. Throughout this paper and during our studies we take the gesture as a collection of signals acquired by different types of sensors from the observed entity. Even though, we can spot a single gesture by searching the starting and ending patterns by the developed gesture recognition methods, when the large number of gestures involved, the performance of the spotting algorithm decreases gradually. However that, if a couple of assumptions are made, such as by the definition, gestures are intentional visible bodily actions which has a start and an end point \cite{mcneill1992hand, kendon2004gesture}, spotting the gesture pattern from a streaming signal problem turns out to be seeking a start and end position of the gestures   instead of seeking it's beginning and ending patterns in a streaming signal. In other words, if the gestures are performed by a hand, there must be some specific hand postures from/to which gestures starts and ends. These positions are represented by the hand postures.

The solution of this research problem can be obtained by simultaneous recognition of the hand posture and gestures that 
reveals the signal pattern boundaries implicitly. In our previous studies we proposed a gesture and posture recognition algorithm and demonstrated commanding a wheelchair by hand postures captured by a Leap motion sensor \cite{boyali2014block, boyali2014hand}. The proposed hand posture recognition method is based on Block-Sparse Representation based Classification (BSRC) which is derived from well-known SRC algorithm families which was first presented in a robust face recognition problem \cite{wright2009robust}. 

In the study \cite{boyali2014block}, the authors extend the SRC based gesture classification methods \cite{boyali2012robust, kavakli2012sign} by assuming block structure on the dictionary matrix and compared the different types of block-sparse solvers for the resulting equations. Achieving a high recognition accuracy with the extended block sparse solvers in \cite{boyali2014block} for all the gesture sets collected by different sensors; 2D Wiimote IR camera and stylus pad, the authors then apply the method for hand posture recognition and demonstrate how to steer a power wheelchair by hand postures captured by a Leap Motion sensor real-time. The application of the BSRC over a long period of time for continuous hand posture recognition yields only a few mis-classifications (up to 0.05\%)  when the hand in transition state between two different postures, indeed while the hand is performing a gesture.

In this study, once the navigation state is activated, the method makes successive decisions top-down to detect whether the hand in a posture, gesture or transition state. At the decision nodes, the candidate hand posture or gesture is detected by Sparse or Collaborative Representation based classification (CRC) algorithms. Unlike the conventional SRC and CRC applications, we build three dictionary matrices for transition, posture and gesture states in which the state representative members are selected from a matrix that contains the features vector as columns. These column vectors are the vectors obtained by a sliding window from the training samples, thus all possible variations are represented in the dictionary and any measurement vector captured by a sliding window from a streaming signal can be labeled if there exist representatives in the dictionary matrix. Although all possible variations of a specific pattern are put by a sliding window in the dictionary, this assumption holds only for the stationary signals that show less variation such as the hand postures which are weak spatio-temporal signals. The difficulty in building such dictionaries for strong spatio-temporal signals such as gestures that come from detecting where the representative sample start in a streaming signal and where it ends. The selected representatives by hand are not of equal length, thereby the resulting system of equations restricts other solution method such as CRC other then SRC. 

In the gesture recognition studies \cite{boyali2014block, boyali2012robust, kavakli2012sign}, a dictionary matrix is constructed from a set of 23 gesture classes, the representative signal samples in which are not of equal length. In these studies, the number of rows of the dictionary matrix is defined by the longest gesture signal that is in a 1D vector form. The gestures with varying shorter lengths are padded with zeros to put the training sample in the dictionary matrix. Padding any vector with zeros does not affect the recovery performance as long as a linear measurement matrix with  Restricted Isometry Property (RIP) is used and the measurement vector can be represented by a sparse linear span of the dictionary columns \cite{candes2006stable, candes2005decoding}. 

In real-time applications in which many signal patterns with varying lengths are to be recognized, a trigger function that alerts the system if the observed signal is represented in the dictionary matrix, must be available to the recognition algorithm. In most of the HCI problems, such kind of triggers might not be readily available. In this application, we acquire the hand position, orientation and velocity measurements by a sliding window with a constant length, build dictionaries for the gestures, turn left, right, go forward, backward, stop; the postures, and transition which is used to detect whether the hand is in a posture or gesture state. If the hand is in a posture state, the variation of the hand motion and orientation values are small but this is not the case in gesturing. 

So far, in the signal pattern recognition studies \cite{boyali2014block, boyaliUn}, we diligently selected the training samples hand by spotting the start and end points of the signal patterns ostensibly to build a discriminating dictionary.  
The success of classification algorithms purely relies on a dictionary which contains the best representatives from each classes to be labeled. On the other hand, the signal is continuously acquired by a sliding window in the real-time gesture recognition algorithm, and it runs after every signal acquisition step. This configuration is not suitable when only a portion of the observed gesture becomes available to the recognition algorithm that reduces the classification accuracy. Both in training and real-time recognition phases, this approach is insufficient for a robust and accurate recognition. We overcome these shortcomings by employing a state of the art Subspace Clustering algorithm (SC) in the training phase to separate the representative samples on the signal without spotting their boundaries. We first perform successive hand gestures by changing the hand from one posture to another, such as in the sequence go forward {$\rightarrow$} turn left {$\rightarrow$} go forward. The obtained training signal encompasses two hand gestures, turn left from go forward position (go{$\rightarrow$}left transition), and returning the beginning posture go forward (left{$\rightarrow$}go transition). The training signal is scanned by a sliding window to build a training matrix in which every measurement is represented as a column vector. If the number of subspaces is two, the clustering algorithm detects two subspaces which generate the gestures such as go-left and left-go in the streaming training signal eliminating error prone human intervention in building gesture dictionary. This approach not only leads a well-discriminating dictionary in which all the phases of a transition gestures are represented and the boundaries are implicitly revealed but also it enables the use of CRC algorithms which yield competently highly accurate at a remarkable computational speed. 

\section{Block-Sparse and Collaborative Representation Algorithms}

Having been introduced in \cite{wright2009robust} robust face recognition algorithms, the SRC methods have reined the signal pattern classification studies for a long time until the block and structured sparse approaches come to the scene. In the SRC robust face recognition study, a sufficient number of face images $v_{k,n_k}$ of different subjects are first down-sampled and put in a dictionary matrix $A=[A_1,A_2,\ldots,A_n]= [v_{1,1},v_{1,2}\ldots,v_{k,n_k}] \in \mathbb{R}^{mxn}$ as a column vector. Then, assuming that any new image $y \in \mathbb{R}^{m}$ belongs to a subject represented in the training dictionary can be well approximated by the linear combination of the training images of the same subject, thus the solution vector $x\in \mathbb{R}^{n}$ in the system of linear equations $y=Ax$ must contain larger coefficients corresponding to the related class location in the dictionary matrix, the coefficients for the unrelated classes take small values become zero. Then, the label $r_{i}$ of the observed image is labeled by finding the minimum reconstruction residual using Eq. \eqref{eq:e1}.

\begin{equation}
\label{eq:e1}
\min_{i}\quad r_{i}(y)=\Arrowvert y-A\delta_{i}(\hat{x_{1}})\Arrowvert_{2}
\end{equation}

In Eq. (~\ref{eq:e1}), $\delta_{i}:\mathbb{R}^{n}\rightarrow\mathbb{R}^{n}$ is the selection operator that selects the coefficients of $i^{th}$ class while keeping other coefficients zero in the solution vector $\hat{x_1}$. The problem setup is similar to that of the Compressed Sensing (CS) research field. In CS, any unknown sparse vector $x$ can be recovered by a random measurement matrix $\Phi \in \mathbb{R}^{mxn}$ with RIP where $m<<n$ by using $\ell_{1}$ regularization.

Due to the fact that, most of the solvers have iterative or alternating structure, thereby the solution takes considerable time for computations. The block or structured sparsity assumption alleviates the computational burden by additionally exposing some form of structure on the sparse vector $x$ to be recovered allowing the block sparse solver pruning out the irrelevant blocks in respect to a threshold value during computations. The authors in the study \cite{boyali2014block} compare  different block sparse solvers with each other and the computation times including that for the conventional SRC for the gesture sets in the \cite{boyali2012robust} and report that although the recognition accuracy slightly improves, the computations are remarkably faster then that of the one used for the SRC in \cite{boyali2012robust}. After the comparisons, the authors opt for the Block Sparse Bayesian Learning (BSBL) solver \cite{zhang2012recovery, zhang2013extension} for the gesture and posture recognition problem in the studies \cite{boyali2014block, boyali2014hand}. 

We tested the Bound Optimization (BO) version of the BSBL solvers family and compared with the Collaborative Representation based classification (CRC) for real time recognition of the braking states with a mobile tablet which is mounted on the top of a Segway mobility robot. Albeit much faster (approximately 20 times) then the conventional SRC based gesture classification on a computer and yielding a higher accuracy, on a brand new mobile device, the computation time for each classification instance takes 0.2 seconds and does not meet real-time requirements \cite{boyaliUn}. 
 
The SRC method affords higher classification accuracy on the grounds that, the method operates on over-complete dictionaries, in which the number of samples from each class is sufficient. In SRC, the observed pattern is coded over these class samples \cite{zhang2012collaborative}. In CRC, the observed pattern is coded over all of the samples in the dictionary, hence the training matrix does not necessarily be over-complete and the method competently gives classification accuracy even the number of samples is insufficient in each of the classes. The general form of the CRC objective function is given in the equation (\ref{eq:e2}) where $\lambda$ is the Lagrange multiplier. 

\begin{equation}
\label{eq:e2}
\min_{x}\quad \hat{x}=\Arrowvert y-Ax\Arrowvert_{q}+\lambda\Arrowvert x \Arrowvert_p
\end{equation}

Depending on the noise level and outliers on the measurements, $p$ and $q$ can take either of $1$ or $2$. When the noise level is low, the  $\ell_{2}$ regularization is the best choice in order to increase the data fidelity \cite{zhang2012collaborative}. In this case the regularized least square objective function becomes;

\begin{equation}
\label{eq:e3}
\min_{x}\quad \hat{x}=\Arrowvert y-Ax\Arrowvert_{2}^{2}+\lambda\Arrowvert x \Arrowvert_2^{2}
\end{equation}

The solution of the optimization problem results in the ridge regression coefficients given in the equation (\ref{eq:e4}).

\begin{equation}
\label{eq:e4}
\hat{x}=({A^T}A+{\lambda}I)^{-1}{A^T}y
\end{equation} 

The different combinations of p and q manifest the variants of the CRC. In the SRC algorithm these values are q = 1 or 2 and p = 1 depending on whether there is a occluded face images in the face recognition applications. The $\ell_{1}$ regularization is more robust then that of $\ell_{2}$ in the representation fidelity in particular when data contain outliers, however, in the absence of outliers, the $\ell_{2}$ regularization yields competent classification accuracy \cite{zhang2012collaborative}.

One of the main advantage that makes the CRC with Least Square Regularization preeminent, the ridge regression operator  $P_{\lambda}=({A^T}A+{\lambda}I)^{-1}{A^T}$ is computed in advance once and stored for the further computations that in turn shortens the computation times beyond comparison with the SRC methods. The pseudo-code of the CRC{\_RLS} method is given in  Table (\ref{table:nonlin}). 

\begin{table}[ht]
	\caption{Pseudo-Codes of the CRC{\_}RLS Method} 
	\centering 
	\begin{tabular}{c l l l} 
		\hline
		
		\hline 
		\hline 
		\rule{0pt}{3ex}  
		1 {\qquad}  & \multicolumn{3}{m{6.5cm}}{Normalize and center the dictionary matrix $A$, center $y$}\\ \hline
		\rule{0pt}{3ex} 
		2 {\qquad}  & \multicolumn{3}{m{6.5cm}}{Code $y$ over $A$ by $\hat{x}=P_{\lambda}y$}           \\ \hline
	    \rule{0pt}{5ex} 
		3 {\qquad}  & \multicolumn{3}{m{6.5cm}}{Label the observed signal by computing the residuals for each of the classes, $Label(y)=\min{r_i}$ where $r_i=\Arrowvert y-A\hat{x} \Arrowvert_2^2$ }                      \\ \hline
		 
	\end{tabular}
	\label{table:nonlin} 
\end{table}

The $\ell_2$ regularization does not promote sparsity but the data fidelity. Therefore, the training samples and the observed pattern must be of equal length that restricts the use of CRC for signal classification problems for which the training class samples show variation in  length. In order to replace the SRC method with the CRC, we defined a constant mean length then re-sampled all the training and test gestures patterns of the study \cite{boyali2012robust}. The recognition accuracy of the CRC is neither more nor less then that of the BSBL\_BO algorithm reported in the study \cite{boyali2014block}. Furthermore, the remarkable computation speeds attained allow the method to be deployed on mobile devices such as mobile phones and tablets for signal pattern classification \cite{boyaliUn}.  

In the real time application of braking state classification study, he signal is acquired by a sliding window as we do in this study. The classification accuracy of the signal pattern acquired by a sliding window is highly dependent on the discriminating power of the training dictionary. In the following section, we describe how to build such a dictionary using a subspace clustering method, when the borders of the signal patterns are indefinite and the neighboring measurements overlaps, due to the sliding window acquisition.  

\section{Implementation}
\subsection{System and Application Architecture}

\begin{figure}[!h]
	\centering
	\includegraphics[scale=0.11]{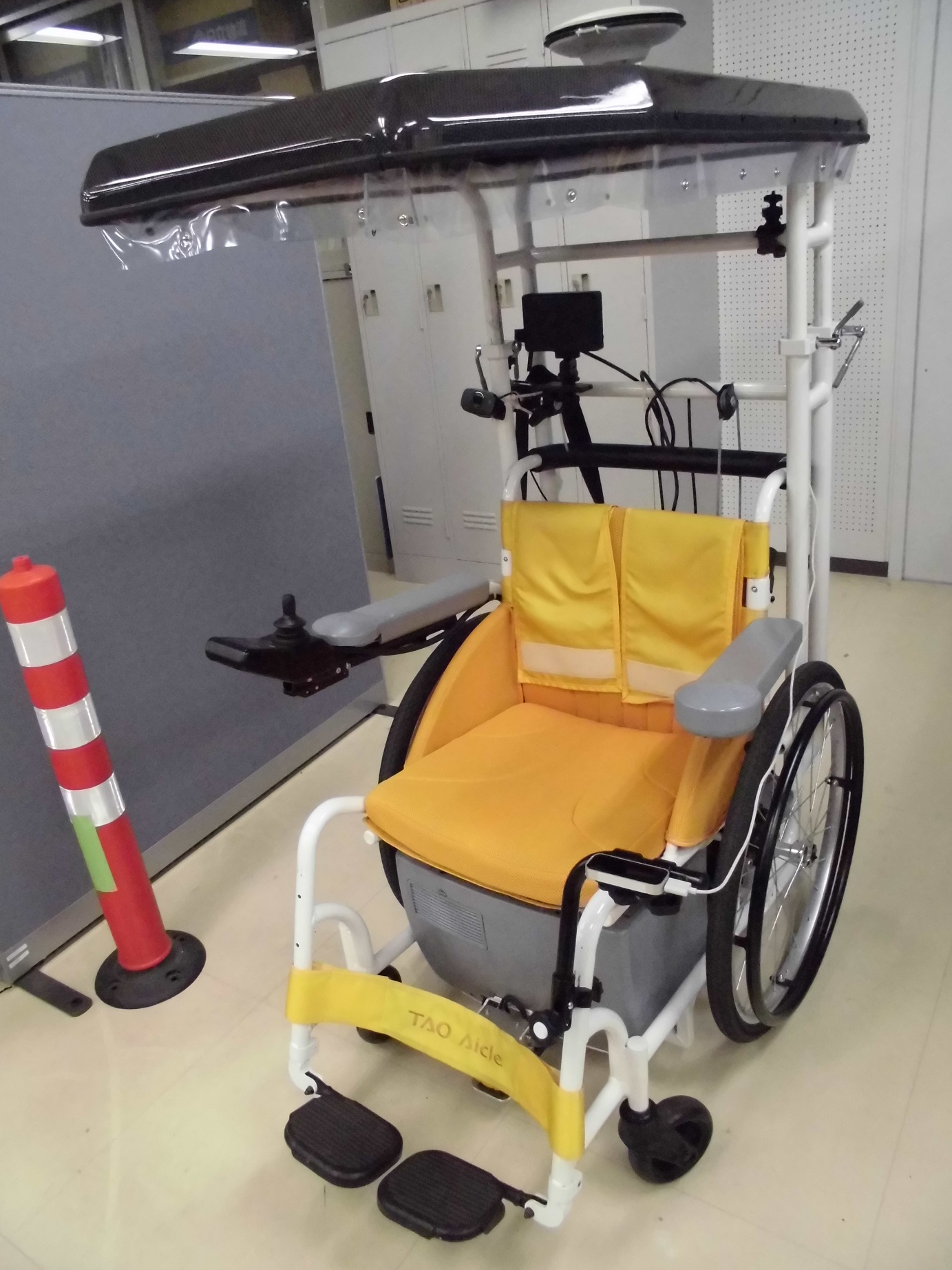}
	\caption{Robotic Wheel-Chair and Leap Motion Sensor}
	\captionsetup{justification=centering}
	\label{fig:f1}
\end{figure}

The robotic wheelchair which has been developed by the Intelligent System Institute of AIST, is equipped with many sensors for localization and mapping in the navigation environment and obstacle avoidance system (Fig. \ref{fig:f1}). 

The wheelchair has the capability of path planning and autonomous modes without human intervention that enable the occupant's hands free. The initial phase of HCI development for this wheelchair, started with the conception to replace the cumbersome emergency button with a more comfortable and smart interface. The most of the wheelchair users are elderly or the people with multiple disabilities whose motor and cognitive skills are highly slowed down in respect to that of able-bodied individuals. Operating an emergency button might be cumbersome in case of emergency for these groups.  As a good candidate in the first phase of the HCI development project, the Leap Motion device is used for this purpose.

\begin{figure}[h]
	\centering
	\includegraphics[scale=0.4]{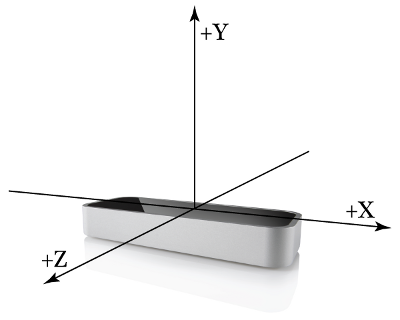}
	\caption{Leap Motion Sensor and Its Coordinate System}
	\captionsetup{justification=centering}	
	\label{fig:f2}
\end{figure}

The Leap Motion is a device (Fig. \ref{fig:f2}) which contains two IR cameras to capture the motion and orientation parameters of a pointer or hands, and reports these measurements approximately at a frequency of 80 Hz. 

The device is placed just under the left arm-support of the robotic wheelchair so that the left hand stays in the effective vision volume of the device. A simple solution for replacing the emergency button with the devices function, is to program the entire system and allowing autonomous mode as long as the device reports a hand within the view of the Leap motion. If a hand is not seen by the device, the autonomous mode is deactivated. 
 
\begin{figure}[h]
	\centering
	\includegraphics[scale=0.35]{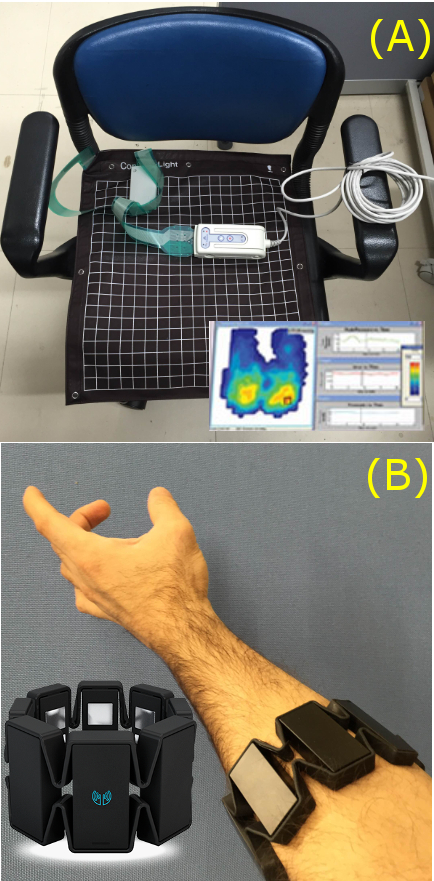}
	\caption{Conformat Pressure Mat and MYO Bracelet }
	\captionsetup{justification=centering}	
	\label{fig:f3}
\end{figure}

This configuration brings additional functionalities to the system allowing the use of hand posture and gesture and to test the developed HCI algorithms with the available sensor systems. It is noteworthy to emphasize at this point that the use of hand posture and gestures for steering a robotic wheelchair is to describe and demonstrate how the recognition algorithm runs real-time but not for suggesting the use of hands for the group of people addressed in the paper. The use of hands for this purpose, requires an eminent hand dexterity. The CRC based signal pattern classification and the training approach we introduce  has hitherto afforded a great success in real-time recognition on various sensor signals such as Wiimote IR, Android tablet motion sensors and Leap Motion device, the authors aim to extend the study with the cutting-edge sensor technologies such as to capture residual body and muscle activities when they become available.  These new technologies such as Thalmic Lab's MYO gesture bracelet and pressure mat (Fig. \ref{fig:f3} B and A) which capture the muscle's electrical activities, motion and position of the worn arm and sitting posture respectively to capture the residual voluntary actions of the addressed group.

\begin{figure}[h]
	\centering
	\includegraphics[scale=0.5]{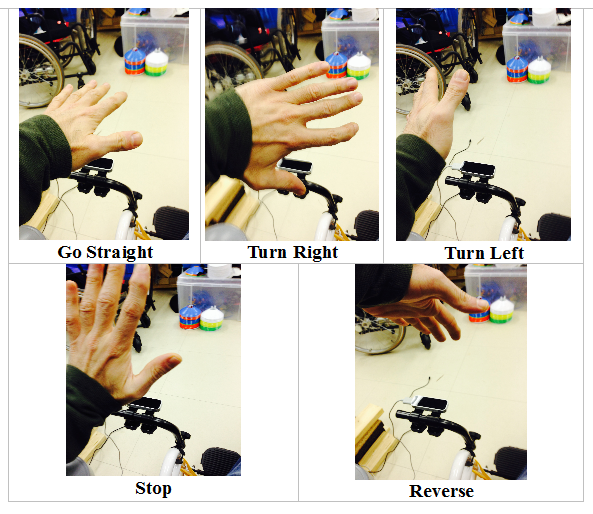}
	\caption{Wheelchair Steering Commands}
	\captionsetup{justification=centering}	
	\label{fig:f4}
\end{figure}

The first party of the MYO devices have been delivered to the developers recently, and the authors have already developed a Matlab interface for this device at the same time when this paper had been indited.  Having sufficiently fine resolution, the pressure mat (Conformat) is capable of sensing the route of changing center of gravity when it is seated on. Tracking the route of center of gravity, the seating posture pressure distribution, the electrical activities of the worn arm are ideally suited and pertinent cues to observe for the people with severe or multiple disabilities. In this regard, tested and simulated classification and training algorithms in different signal domains lay a foundation out for multi-modal analysis and synthesis. Five steering commands are defined for wheelchair navigation; Go Straight, Turn Left, Turn Right, Stop and Go Backward (reverse). The hand postures for these commands are illustrated in Fig. \ref{fig:f4}. 

\begin{figure}[h]
	\centering
	\includegraphics[width=\columnwidth]{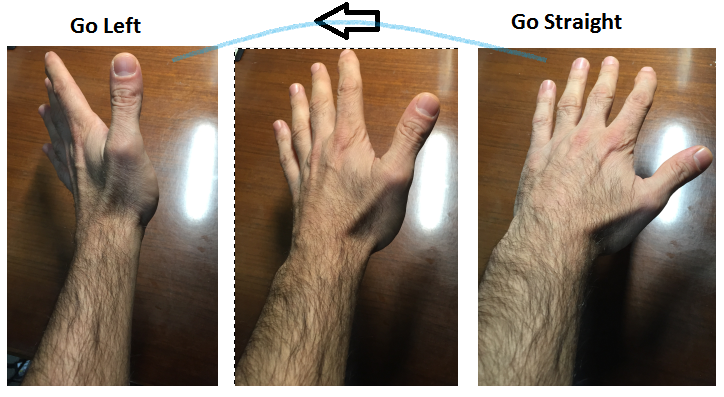}
	\caption{Go Left Transition Between Go and Left Postures}
	\captionsetup{justification=centering}	
	\label{fig:f5}
\end{figure}

Owing to the fact that, in the real-time application, the hand does a gesture while changing it's one posture to another. We name these transition or gestural states which are in total eight as, Go2Left (Fig. \ref{fig:f5}), Left2Go, Go2Right, Right2Go, Go2Stop, Stop2Go and Go2Reverse, Reverse2Go.  

The classification algorithm has no limitation to introduce a new posture and gesture command into the training dictionary, thus, more functionalities can be devised such as if required increasing or decreasing the speed of the wheelchair. 

\begin{figure}[h]
	\centering
	\includegraphics[width=\columnwidth]{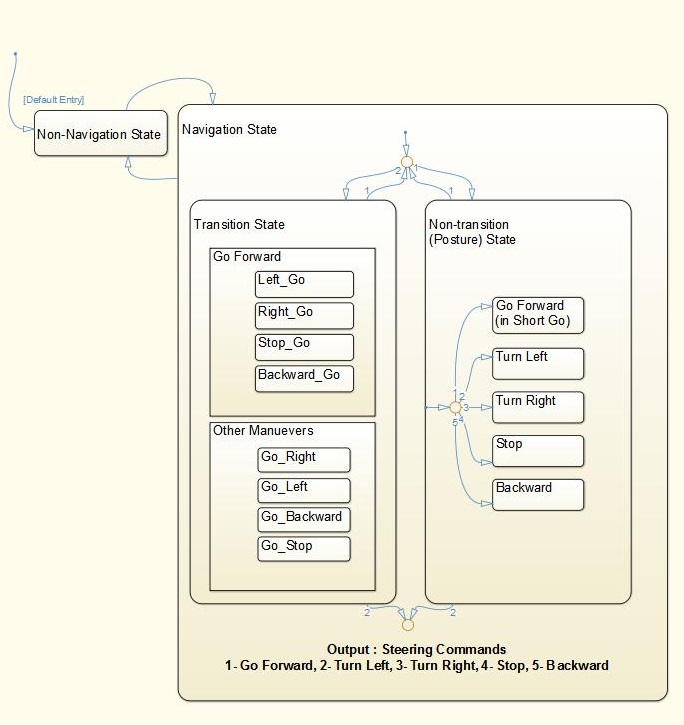}
	\caption{Matlab Stateflow Diagram of the Classification Scheme}
	\captionsetup{justification=centering}	
	\label{fig:f6}
\end{figure}

The overall classification scheme is given in Fig. \ref{fig:f6} which shows the Finite State Diagram created in the Matlab Stateflow environment. When a hand seen by the Leap Motion device, the system switches into the navigation mode. The classification application first looks at whether the detected hand is in a transition or doing posture and enter into one of the either states after the first decision is made. The first level transition dictionary is used at the first decision node. The transition matrix is composed of 200 samples from the posture and gesture states in total the number of columns of the first dictionary is 400. The group of posture, non-transition states is composed by 40 samples from each states. Whereas there are eight gestures and we only need 25 representative from each of the eight gestures. The most simple and discriminating feature is the magnitude of palm velocity vector for detecting whether the hand is doing a posture or gesture, thereof is used. 
 
If the algorithm enters into the posture state, the classification method of the application; CRC or Block Sparse BSBL\_BO, selects the most probable hand posture state out of five by computing the minimum representation residuals. The determined class label is then sent to the on-board wheelchair computer.

Eight hand gestures which are embedded in the four hidden bilateral transitions, are possible in the transition block. Every hidden state is the base of two hand gestures which are deciphered in the training phase. The underlying rationale of hidden transition states which are composed of successive bilateral gestures such as Go2Left - Left2Go or Go2Stop - Stop2Go that the gesture couples are performed successively in gathering training signals to isolate the neighboring gestures or the transitions from others for building clean dictionaries. The gesture couples are deciphered in the training phase by a clustering algorithm which is detailed in the following subsection.

\begin{figure}[h]
	\centering
	\includegraphics[width=\columnwidth]{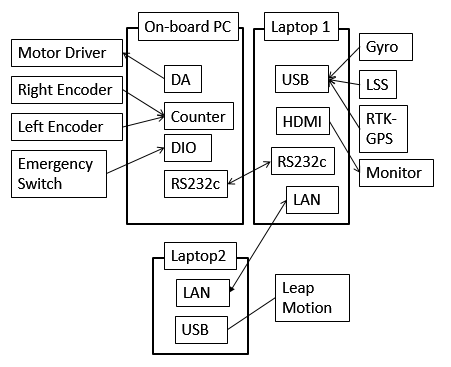}
	\caption{Hardware Architecture of Entire System}
	\captionsetup{justification=centering}	
	\label{fig:f7}
\end{figure}

On the center of each gesture state is the Go Straight hand posture. In every transition, the hand visits Go Straight posture. The reciprocal transitions between other states such as the transition between the Turn Left to Turn Right hand postures are not included in the system, albeit possible.  Every transition leaving a hand posture state is assigned to the new steering command that the gesture state ends up. For instance, if the hand is gesturing and the gesture recognition algorithm labels the gesture as Go2Left indicates that the hand was on the Go Straight hand posture and now it is going to Turn Left one, thus the system assigns the Turn Left steering command to the gesture transition leaving the Go Straight hand posture \ref{fig:f5}.  The labeled class is then assigned to a steering command that is sent to the on-board computer of the wheelchair from a laptop over a network cable (Fig. \ref{fig:f7}). 
 
The hardware structure of the robotic wheelchair is given in Fig. \ref{fig:f7}. In this configuration, an on-board computer gathers the sensor measurements from the wheels, emergency switch and control the electric motors of the wheelchair. 

A second laptop, on which the most of the robotic tasks are performed communicates with the on-board computer over a RS232 protocol. A motion sensors, GPS and laser scanner for mapping the environment and obstacle avoidance are connected to an external computer which hosts the application for robotic functionalities. The 3rd laptop is only for developing HCI applications and It is only connected to the robotic wheelchair to test the classification algorithms real-time before writing the interface application and deployment on the Laptop (1). 

\subsection{Training}

The CRC and SRC based signal pattern recognition require a training dictionary in which the representative samples from each classes to be recognized. Depending on which method is utilized, the number of samples from each class might show variation that exerts influence to the classification accuracy. It is fairly easy to build a dictionary for the application on which the boundary of the signal patterns are well defined.  For instance, in the study \cite{boyali2012robust}, each Wiimote gesture is repeated 15 times, 2/3 of each set is reserved for training and the remaining for testing the recognition accuracy. The 2D gestures are performed by hand holding an IR pen that is in view of the Wiimote's IR camera. However while performing gestures, the performer turn on the IR LED on the tip of the pen by pressing a button, hence the Wiimote IR camera only capture the visible IR light when the LED is turned on and gesture's start and end points are well defined. In case of lack of such a boundary defining tool analogous to the button switch on the IR pen, one can define the signal pattern boundaries by pinpointing visually on the signal plot. This procedure requires diligent and tedious efforts and is always susceptible to selecting ambiguous signal boundaries. Additionally, determining the boundaries by hand  might result in the dictionaries in which the signal patterns overlap reducing the discrimination power of the classification system. 
 
\begin{figure}[h]
	\centering
	\includegraphics[width=\columnwidth]{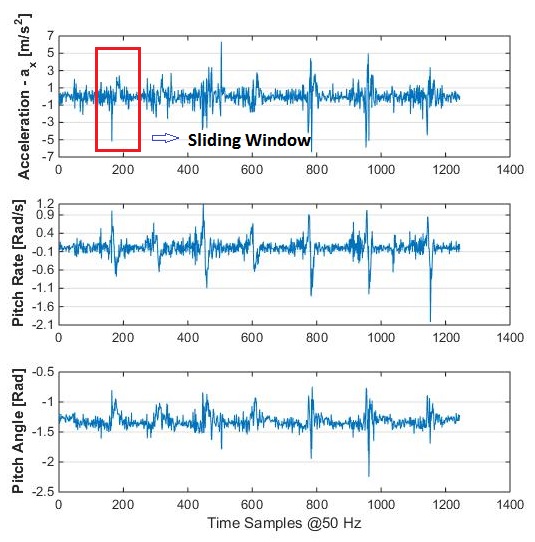}
	\caption{Hardware Architecture of Entire System}
	\captionsetup{justification=centering}	
	\label{fig:f8}
\end{figure}

On the other hand, the training samples even in the same class might exhibit variation in length such as in the braking state classification of the Segway study \cite{boyaliUn}. In this study when the rider of Segway performs a normal stop, the boundaries of this breaking maneuver on some signal features cannot be distinguished visually especially on the noisy signals. It necessitates to inspect other features to seek the pattern boundaries. This phenomena is shown in Fig \ref{fig:f8}. 

The figure shows the measurements from an Android device attached on the top of a Segway's handle bar. The rider does a periodic normal breaking which is described as the braking maneuver, is slow and a common pattern when riding on public roads. The motion direction related features for the test setup are the longitudinal acceleration, the angular velocity and pitch rotation of the Segway's handle bar around the $y$ axis. The test signal exhibits periodicity that can be seen on the longitudinal acceleration and the pitch angle, however the gyroscope readings at the bottom in Fig. \ref{fig:f8} does not exhibit explicit periodicity and one cannot spot the boundaries. Besides the measurements are highly noisy which complicate isolating the pattern. In this training signal, two states are available; normal braking pattern and cruising that includes acceleration. We pick the signal boundaries hand and do not use any clustering algorithm which brings about additional flexibility and robustness in the real-time signal pattern classification applications. 

\begin{figure}[h]
	\centering
	\includegraphics[scale=0.4]{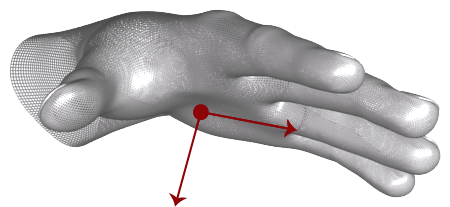}
	\caption{Unit Palm Normal Vector \cite{leapon}}
	\captionsetup{justification=centering}	
	\label{fig:f9}
\end{figure}

A sliding window captures the signal at a specified frequency in the real-time applications. For this type of signal acquisition, the captured signal pattern might not be represented in dictionary and  might combine  smaller portions of any two signal patterns represented in the dictionary. This phenomenon is also undesired for the continuous signal pattern classification. We adopt a clustering algorithm in the training phase to mitigate proclaimed complications without leaving a gap between the captured signal in the real time application and the training dictionary. The representative samples are collected in the same manner as in the real time application by a sliding window then the captured samples are put in a dictionary matrix. For example, a hand gesture training signal are recorded for 20 seconds at a frequency of 50 Hz resulting in 1000 measurements in the training signal. The features for the posture and gesture recognition are the palm normal (Fig. \ref{fig:f9}) represented with a unit vector and the orientation angles of the palm; roll, pitch and yaw.  

\begin{figure}[h]
	\centering
	\includegraphics[width=\columnwidth]{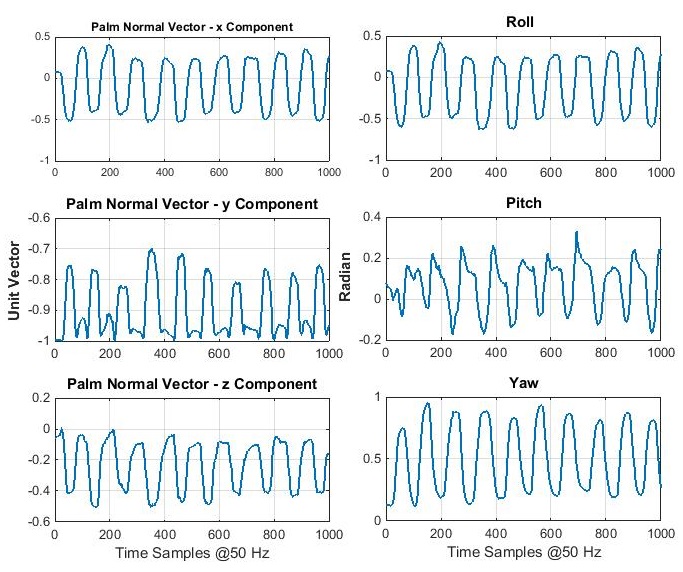}
	\caption{Go Straight - Turn Left - Go Straight Training Features}
	\captionsetup{justification=centering}	
	\label{fig:f10}
\end{figure}

These features (Fig. \ref{fig:f10}) are converted the 1D vector at every time instance to put the sample as column vector in the training matrix. The width of the sliding window is 25 measurements which is sufficient to represent a gesture, since a transition takes place approximately a half second which corresponds to 25 time samples. The length of the column vector thus become six window widths because of that the six features are used for the signal patterns.

One of the great merits that Sparse Representation and the CS research field have inspired the subspace clustering methods which assumes the high dimensional data are generated by a union of subspaces that lies on a low dimensional manifold \cite{elhamifar2013sparse}.  Being heaped up around two pivoting approaches, Low Rank Representation (LRR)  \cite{liu2013robust} and Sparse Subspace Clustering (SSC) \cite{elhamifar2013sparse}, these techniques have gained reputation over abound application areas where big data is involved. The subspace clustering methods differ in the objective function used for the optimization problem, noise tolerance, outlier detection and the penalty functions incurred. 

The LRR method pose the subspace clustering as a minimum rank problem in which the nuclear norm is minimized. By virtue of the nuclear norm, the resulting objective function is a convex optimization problem and many different solution methods can be used. The general form of the objective function of the LRR method is given in Eq (\ref{eq:e5}).

\begin{equation}
\label{eq:e5}
\min_{Z}\quad  \Arrowvert Z\Arrowvert_{*} \quad s.t. \quad X= AZ+E
\end{equation}

In Eq. (\ref{eq:e5}), $X$ represents the observed data, $A$ and $Z$ the matrix of union of subspaces or dictionary and the coefficient matrix which is to be minimized respectively. The error term is represented by $E$ in the equation. 

In a diverse range of application fields, the observed data matrix $X$ are used as the dictionary as in the SRC method. Thus, the objective function in these applications becomes;

\begin{equation}
\label{eq:e6}
\min_{Z}\quad  \Arrowvert Z\Arrowvert_{*} \quad s.t. \quad X= XZ+E
\end{equation}

The SSC method makes a concrete assumption, similar to one made in the LRR applications, any column vector in the observed data matrix $X$ can be represented by the linear combination of the available column vectors excluding the represented vector. Therefore, the method is based on self-expressiveness property. The objective function of the SSC is given in Eq (\ref{eq:e7}). 

\begin{equation}
\label{eq:e7}
\min_{Z}\quad  \Arrowvert Z\Arrowvert_{1} \quad s.t. \quad X= XZ+E, \quad diag(Z)=0
\end{equation}
 
The sparsity promoting norm $\ell_{1}$ is used in the objective function. An affinity matrix is composed by various methods using the solution matrix $Z$ in both of the methods. The resulting affinity matrix are segmented into the subspace groups by a spectral clustering algorithm Normalized Cuts (Ncut) \cite{shi2000normalized}.

We made use of a new subspace clustering algorithm derived from the SSC approach; Ordered Subspace Clustering (OSC) \cite{tierney2014subspace}  which incures a new penalty to the objective function. Thereby, the new term reveals the neighboring information in the data matrix by imposing column similarity to the objective function. The objective function of the OSC method is as follows;  

\begin{align*}
\label{eq:e7a}
\min_{Z, E}\quad \frac{1}{2} \Arrowvert E\Arrowvert_{F}^2+\lambda_{1}\Arrowvert Z\Arrowvert_{1} +\lambda_{2}\Arrowvert ZR\Arrowvert_{1,2}\\
\quad s.t. \quad X= XZ+E  
\end{align*}

where $R$ in the equation is a lower bidiagonal matrix, all the elements in the main and lower diagonal of which are respectively -1 and 1. 

\[ 
 R =
  \begin{bmatrix}
   -1 &&&&&  \\
   1 & -1 &&&&  \\
    & 1&-1  &&& \\  
    && \ddots&\ddots&& \\  
    &&&1&-1
  \end{bmatrix}
\] 
 
The second penalty term $\Arrowvert ZR\Arrowvert_{1,2}$ enforce neighboring columns of coefficient matrix $Z$ to be similar such that $z_{i}\thickapprox z_{i+1}$.

The nature of our training and real-time classification problem, sequential classification with a dictionary that is constructed in the same manner, the OSC problem settings are completely in alignment. The use of a clustering algorithm fits the sequential pattern classification problem eases to build a dictionary with a powerful discrimination capability. As such, it predominates through out the operation of the classification application for robustness and the accuracy.

The OSC admits the training matrix $X \in \mathbb{R}^{150x975}$ for each of the gesture couple training signals and label the columns of the matrix in association with their affinity.  We practically define the number of union of subspaces as two for training gestural transitions. The number of columns labeled for each of the subspace is approximately the same. We choose 100 columns from the each subspace randomly and build two dictionaries which are then to be concatenated after repeating the same procedures for all the transitions. We use this training perspective for only the gestural transitions between the postures. The signal is almost stationary for the postures and there is no need to use clustering as only one posture is recorded, thus one cluster is available. 

\section{Simulations and Results}

In our previous study \cite{boyali2014hand}, we demonstrated steering a robotic wheelchair with only the hand postures. The BSBL\_BO method is employed for real-time classification. The postures in previous and this study assigned to the steering commands are given in the Table \ref{table:table2}. The eight gesture transition are also included in the table. 

\begin{table}[ht]
	\caption{Steering Commands Assigned to Gestures and Postures} 
	\centering 
	\begin{tabular}{c l l } 
		\hline
		
		\hline 
		\hline 
		\rule{0pt}{3ex}  
		Command Value {\qquad}  & \multicolumn{2}{m{5.5cm}}{Assigned Gestures and Postures}\\ \hline
				
		\rule{0pt}{3ex}  
		1 {\qquad}  & \multicolumn{2}{m{5.5cm}}{Posture: Go Straight, Gestures; Left2Go, Right2Go, Stop2Go, Reverse2Go }\\ \hline
		\rule{0pt}{3ex} 
		2 {\qquad}  & \multicolumn{2}{m{5.5cm}}{Posture: Turn Left, Gesture:Go2Left }\\ \hline
	    \rule{0pt}{3ex} 
		3 {\qquad}  & \multicolumn{2}{m{5.5cm}}{Posture: Turn Right, Gesture: Go2Right} \\ \hline
		\rule{0pt}{3ex} 
		4 {\qquad}  & \multicolumn{2}{m{5.5cm}}{Posture: Stop, Gesture: Go2Stop} \\ \hline
		\rule{0pt}{3ex} 
		5 {\qquad}  & \multicolumn{2}{m{5.5cm}}{Posture: Reverse, (Backward), Gesture: Go2Reverse} \\ \hline
		\rule{0pt}{3ex} 		 
	\end{tabular}
	\label{table:table2} 
\end{table}

\begin{figure}[h]
	\centering
	\includegraphics[width=\columnwidth]{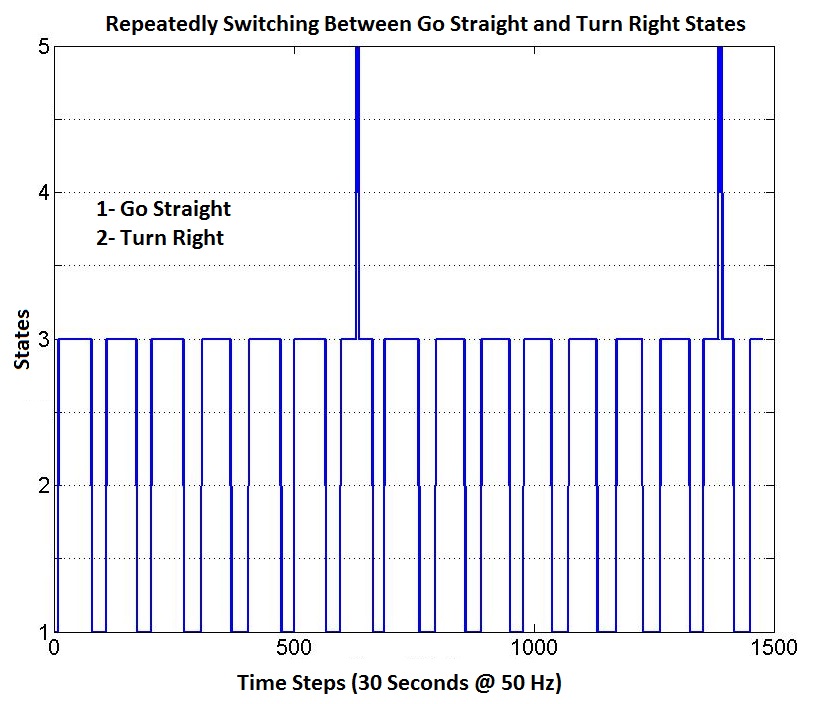}
	\caption{Go Straight - Turn Right Steering Commands}
	\captionsetup{justification=centering}	
	\label{fig:f11}
\end{figure}

In the study \cite{boyali2014hand}, the classification algorithm yields only a couple of false recognition out of thousands. The spikes seen in Fig. \ref{fig:f11} shows these mis-classifications which are only two instances in the 1500 successive recognition results. 

These mis-classification spikes are eliminated during the real-time with a simple filter similar to moving average. The application stores the last five steering commands in a filter and finds the outlier and decides the most probable steering command. The spikes in Fig. \ref{fig:f11} stems from the transition states the hand visits while changing the position one postures to another.  

We developed a signal pattern classification algorithm to detect braking states of a Segway. In this study \cite{boyaliUn}, since the classification application runs on an Android tablet which is mounted on the top of the handle bar of a Segway, although the BSBL method gives a very high accuracy for the real-time application, the computation time at every time instant takes approximately 0.2 seconds which is not a desired computation time for the real time application. 

We replaced the BSBL framework with the CRC method and the computation times for each instant have tremendously improved without compromising the accuracy. Seeing that, we added the gesture recognition module, introduced a new training perspective and used the CRC for signal pattern recognition in this study over the only posture based interface. 

\begin{figure}[h]
	\centering
	\includegraphics[width=\columnwidth]{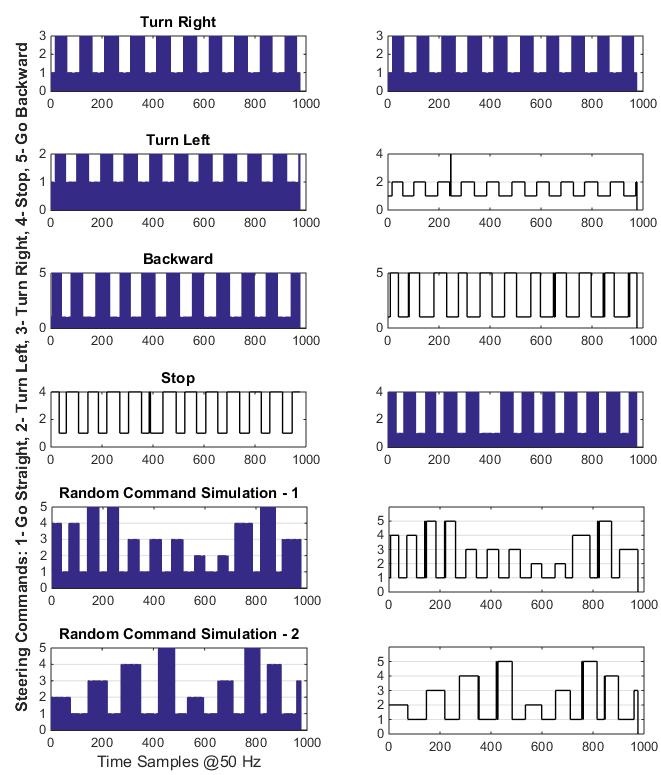}
	\caption{Only Gesture and Combined Gesture and Posture Recognition Simulations}
	\captionsetup{justification=centering}	
	\label{fig:f12}
\end{figure}

The results are very promising. We included Fig. \ref{fig:f12} for comparison purpose and how the training perspective improves for the previous posture recognition problem setup.

The Fig. \ref{fig:f12} shows the recognition results of the BSBL algorithm. The first column shows the steering command for various hand gesture simulations, whereas the second is for combined gesture and posture classification. On the top, four rows are the simulation results while hand change position only two hand positions that are Go2Right, Go2Left, Go2Backward and Go2Stop gestural transitions. The remaining rows are the simulations for random hand gestures in which the hand visits every hand postures defined in the application. The sub-figures that contain mis-classification are plotted by the stairs command of Matlab, if there is no mis-classification the figures are plotted by a solid color bar. Every simulation outputs 975 of recognition label. As seen in the figure there is only one misclassification instance in the first column for gestural transition. The application mis-classify a posture. When a combined version is used, the application corrects this error as seen in the second column for Go-Stop simulation. It is interesting to note that, the mis-classifications only occurs between the neighboring transitions because of the decisition tree structure of the application. In the Fig. \ref{fig:f11}, the misclassified signal patterns show the hand posture state as Stop while the hand switching between Right and Go Straight postures. The Stop posture is not a neighbor to this state. Errors only occurs between two neighboring states even they are rare. 

In the BSBL setting, if the combined gesture and posture recognition approach, the simulation results of which in the second column in the figure, this phenomenon suggests that only the posture recognition states produce more misclassification but manageable than that of gesture but less then the reported ratio in the previous study. The error producing state in the posture classification is the Go Backward hand state as seen in the figure group of second column. There are a few mis-classifications between the Go Backward postures when the hands stay either of these states in all the simulations Go Backward hand posture is visited.

\begin{figure}[h]
	\centering
	\includegraphics[width=\columnwidth]{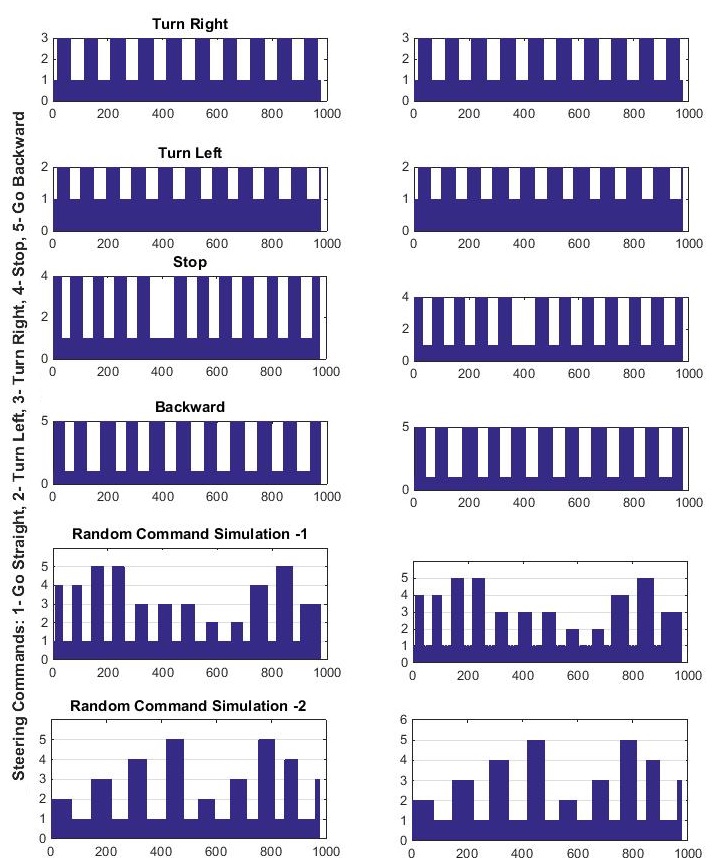}
	\caption{Only Gesture and Combined Gesture and Posture Recognition Simulations}
	\captionsetup{justification=centering}	
	\label{fig:f13}
\end{figure}

The picture described above completely changes when the CRC is used for classification on the same problem setup with the same dictionaries. The results for the 12 simulations are given in Fig. \ref{fig:f13} for CRC classification.  

There is no any single mis-classification exists in the simulations. The only gesture, posture and combined versions yield a perfect accuracy as shown in the figure. The computation times are unequivocally are beyond comparison because of the pre-computed projection operator in the CRC algorithm and iterative structure in the sparse solver. The computation times when the CRC and BSBL methods are used for 975 instances of labeling are 112 and 0.39 seconds respectively. 

\begin{figure}[h]
	\centering
	\includegraphics[width=\columnwidth]{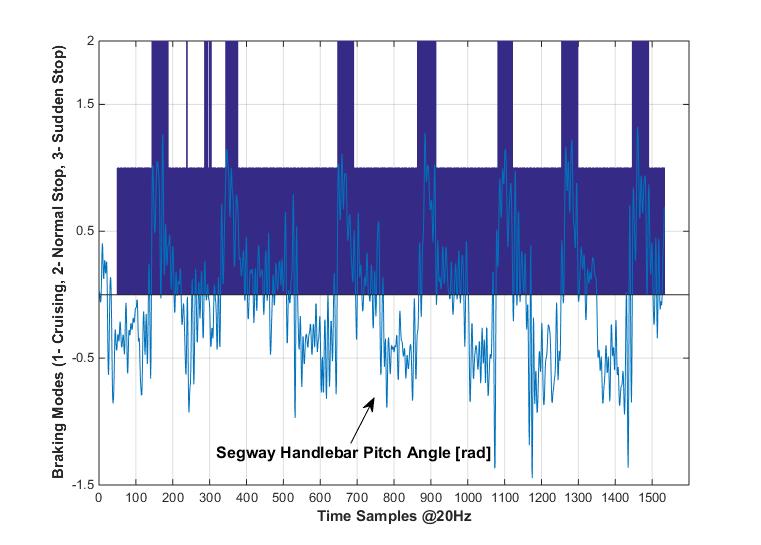}
	\caption{Normal Braking Experiment}
	\captionsetup{justification=centering}	
	\label{fig:f14}
\end{figure}

We implemented the same training and signal pattern classification framework also to detect the braking states of a traveling Segway \cite{boyaliUn}. The Fig. \ref{fig:f14} and \ref{fig:f14} show the results of the real-time braking mode classification. Three classes are represented in the dictionary which are the cruising (no braking), normal and sudden braking. 

\begin{figure}[h]
	\centering
	\includegraphics[width=\columnwidth]{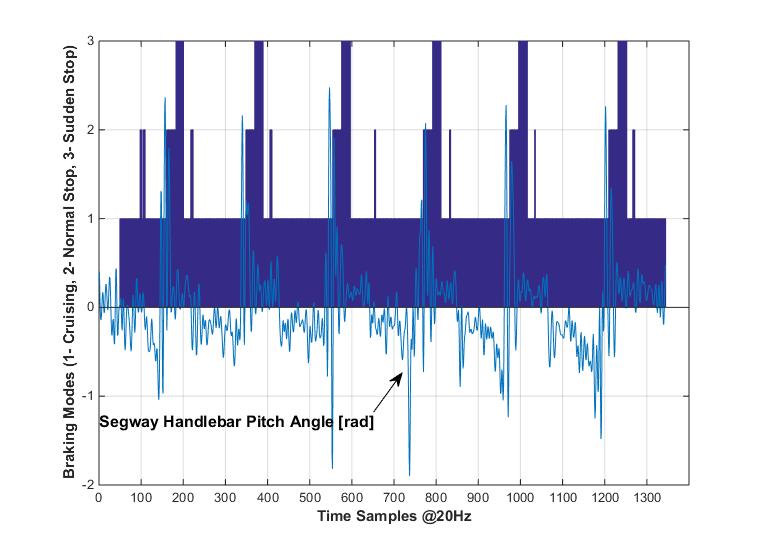}
	\caption{Sudden Braking Experiment}
	\captionsetup{justification=centering}	
	\label{fig:f15}
\end{figure}

In the normal (Fig. \ref{fig:f14}) and sudden braking (Fig. \ref{fig:f15}) experiments, the rider repeatedly accelerates, reach a cruising speed and perform normal or sudden braking maneuvers. As seen in the figures, in the normal breaking experiment there is no sudden breaking label whereas the sudden breaking results contain normal breaking modes. This is due to that every sudden breaking maneuver contains normal breaking phase and the algorithm catches these phases.  
   
\section{Conclusion and Future Works}

In this study, we fathom the methodologies and the framework for developing intuitive HCIs and demonstrate the use of state of art signal pattern classification methods for sequential gesture and posture recognition of a detected hand by a Leap motion device. The algorithms and approaches detailed here also culminated in different user interfaces for the difference sensor domain and application requirements such as the Segway braking mode classification. 

The framework we draw on here, not only leads to overwhelmingly accurate classification regardless of the sensor domain, but also, it unveils the class of the observed signal pattern sequentially by a decision tree regardless of whether the signal patterns vary in length or not. It is also closely related to signal pattern spotting on a streaming signal due to the signal used in training process. The signal used for training encompasses two different gesture patterns, the boundaries of which are not known in advance. Therefore, the boundary information is implicitly decoded through out the decision tree.

The authors will extend the study with a pressure mat which can capture seating posture from the pressure distribution and track the center of the gravity of the occupant, and the recent device Thalmic Lab's gesture bracelet MYO which can report the electrical activities of the worn arm wireless to create a multi-modal HCI for the people with severe impairments or multiple disability. 

The training phase for power wheelchairs is not an easy process for most of the prescribed patients as stated in the introduction. For this reason, we will start development of a virtual reality training simulator run on mobile platforms  integrating multi-modal HCIs. 

\section*{Acknowledgment}

The first author acknowledges the support of Japan Society for the Promotion of Science fellowship program.

\ifCLASSOPTIONcaptionsoff
  \newpage
\fi



%

\bibliographystyle{IEEEtran}
\bibliography{IEEEfull,thebib}

%

\begin{IEEEbiographynophoto}{Ali Boyali} 
Dr. Ali Boyali is a post doctoral research fellow at the Intelligent Systems Institute of AIST supported by the Japan Society for the Promotion of Science (JSPS) fellowship program. He received his BSc, MSc and PhD degrees from Istanbul Technical University in Mechanical and Automotive Engineering. He researched gesture and posture recognition systems and interfaces for virtual reality applications at Macquarie University as a post doctoral researcher between 2010-2012. His current researches are on Intuitive Human Computer Interfaces for the mobility devices. 
\end{IEEEbiographynophoto}
 
\begin{IEEEbiographynophoto}{Naohisa Hashimoto}
Naohisa Hashimoto is a senior researcher of the Smart Mobility Research group at the National Institute of Advanced Industrial Science and Technology (AIST) in Japan. He received Ph.D. degree from the Keio University in 2005. During 2010-2011, he was a visiting researcher at the Ohio State University and Centre of Automotive Research
\end{IEEEbiographynophoto}
 
 \begin{IEEEbiographynophoto}{Manolya Kavakli}
Associate Professor Manolya Kavakli is currently the Director of the Postgraduate Coursework Program at the Department of Computing, Macquarie University. She is an active researcher working on Human Computer Interaction (HCI) in the last 25 years with 133 refereed papers (775 citations).She has been the recipient of 10 awards and more than 30 grants from a number of scientific international organisations including the Australian Research Council, the Scientific and Technical Research Council of Turkey. 
 \end{IEEEbiographynophoto}




\end{document}